%% file: 0main.tex
\documentclass{article} 
\usepackage{iclr2027_conference,times}

\input{math_commands.tex}

\usepackage{hyperref}
\usepackage{url}
\usepackage{enumitem}
\usepackage{booktabs}
\usepackage{multirow}
\usepackage{makecell}
\usepackage[table]{xcolor}
\usepackage{graphicx}
\usepackage{placeins}
\usepackage{caption}
\usepackage{subcaption}
\usepackage{CJKutf8}
\usepackage{xcolor}
\usepackage{float}

\title{Frame Differential On-Policy Self-Distillation for Video Reasoning}

\author{
Haiying He\textsuperscript{1},
Xin Zheng\textsuperscript{1},
Shaoli Hu\textsuperscript{1},
Shijun Xiao\textsuperscript{2},
Xuanhe Liu\textsuperscript{3},
Bing Li\textsuperscript{4},
Harry Yang\textsuperscript{1$\dagger$}
\\[2mm]
\textsuperscript{1}HKUST
\quad
\textsuperscript{2}NKU
\quad
\textsuperscript{3}SEU
\quad
\textsuperscript{4}KAUST
\quad
\\[2mm]
}

\newcommand{\methodshort}{\textsc{FD-OPSD}}

\iclrfinalcopy 
\begin{document}

\maketitle
\begingroup
\renewcommand{\thefootnote}{}
\footnotetext{
\textsuperscript{$\dagger$} Corresponding author. 
}
\addtocounter{footnote}{-1}
\endgroup
\begin{abstract}
Reinforcement learning (RL) has substantially improved the reasoning ability of multimodal language models through verifiable rewards and increasingly fine-grainedvisual or temporal credit assignment. In video reasoning, however, current RL methods typically train with a fixed sparse frame budget: increasing the number of frames makes autoregressive rollouts expensive, while too few frames may miss temporally localized events and fine-grained visual details. We present \textbf{Frame Differential On-Policy Self-Distillation (\methodshort{})}, which transfers the useful evidence of dense frame observations to a sparse frame policy during RL training. \methodshort{} compares the policy's token level preferences for the same sampled response under sparse and dense views, and distills the resulting frame differential signal without an external teacher or dense autoregressive rollout. The method preserves sparse-frame rollouts and leaves inference unchanged. Across Qwen2.5-VL-7B and Qwen3-VL-4B on six video reasoning benchmarks, \methodshort{} yields higher overall average performance than the strongest corresponding GRPO, T-GRPO, or Video-KTR baselines across the 16, 32, and 64 frame evaluation settings. These results show that dense visual evidence can be transferred selectively during training through token level self-distillation while retaining sparse frame rollouts and unchanged inference. 
Code is available at \url{https://github.com/wannanfeng/video_reasoning}.

\end{abstract}

\input{1introduction}
\input{2relatedwork}
\input{3method}
\input{4experiment}
\input{5conclusion}
\input{6limilation}

\newpage
\bibliographystyle{iclr2027_conference}
\bibliography{iclr2027_conference}

\newpage
\appendix
\input{appendix}
\end{document}

%% file: math_commands.tex
\usepackage{amsmath,amsfonts,bm}

\def\eqref#1{equation~\ref{#1}}

\def\1{\bm{1}}

\DeclareMathAlphabet{\mathsfit}{\encodingdefault}{\sfdefault}{m}{sl}
\SetMathAlphabet{\mathsfit}{bold}{\encodingdefault}{\sfdefault}{bx}{n}



%% file: 1introduction.tex
\section{Introduction}
Reinforcement learning (RL) has become an effective way to improve the reasoning abilities of large language models and multimodal language models~\cite{adar1,kimi,DEEPSEEKR1}. Its success has been driven by verifiable outcome rewards, group relative policy optimization~\cite{grpo,dapo}, and task specific reward~\cite{videochat,visual_RFT} or credit assignment design~\cite{GIGPO}. In video reasoning, GRPO style policy optimization has provided a practical route to improve multimodal reasoning from outcome rewards. Video-R1~\cite{videor1} introduces T-GRPO to encourage sensitivity to frame order, while Video-KTR~\cite{videoktr} performs modality aware token level policy shaping. These methods demonstrate that the design of rewards and policy updates is central to video reasoning, but they generally train with a fixed frame sampling budget.

The frame budget creates a distinct difficulty for video RL. More frames can expose events and details that are absent from a sparse view~\cite{RL2LONGVIDEO}, yet the cost is amplified during autoregressive rollouts, where multiple responses are generated for each prompt. Sparse training views can therefore miss temporally localized evidence~\cite{Framemind,ThinkingINVIDEO}, while direct dense frame training increases rollout cost and introduces substantial redundant or irrelevant visual context~\cite{LONGVIDEOR1}. This leads to a question that is largely orthogonal to reward design: \textbf{\textit{Can dense frame evidence improve sparse frame reasoning without requiring dense frame rollouts?}}

To address this frame budget mismatch, we propose \textbf{Frame Differential On-Policy Self-Distillation (\methodshort{})}. \methodshort{} keeps the policy rollout sparse, but uses a denser view of the same video to provide an auxiliary training signal for the sampled response. By comparing the policy's token level preferences under sparse and dense views, the method identifies response positions whose predictions change when additional temporal evidence is available. This frame differential signal is selectively transferred to the sparse view policy through a confidence aware fidelity objective while retaining the underlying GRPO optimization. The dense view is used only for token scoring during training, so \methodshort{} requires neither a separately trained teacher nor a dense autoregressive rollout and does not change inference cost.

Across Qwen2.5-VL-7B~\cite{qwen2vl} and Qwen3-VL-4B~\cite{qwen3vl}, six video reasoning benchmarks, and 16, 32, and 64 frame evaluation settings, \methodshort{} achieves higher overall average performance than the strongest corresponding GRPO, T-GRPO, or Video-KTR baseline. Additional token-level analysis shows that the frame differential signal is concentrated on video semantic content, including temporal expressions, actions, and visual entities. Together, these results support dense evidence transfer as a complementary direction to reward and policy design for efficient video RL. To summarize, our contributions are:
\begin{itemize}[leftmargin=*]
\item We identify the sparse training frame budget as an underused source of supervision in video RL and formulate dense frame assistance as an evidence transfer problem.
\item We introduce \methodshort{}, a frame differential distillation method that transfers dense view evidence to sparse frame RL without a separate teacher or dense decoding.
\item We demonstrate overall gains across two Qwen-VL backbones, six video reasoning benchmarks, and three evaluation frame budgets, together with token level evidence that clarifies where the transferred signal is concentrated.
\end{itemize}

%% file: 2relatedwork.tex
\section{Related Work}
\subsection{Reinforcement Learning for Large Language Models}

Reinforcement learning has become a widely used post-training paradigm for improving large language models beyond supervised imitation. Preference based alignment and, more recently, reinforcement learning with verifiable rewards~\cite{instructgpt,tulu3} (RLVR) have enabled substantial progress on reasoning tasks whose outcomes can be evaluated automatically~\cite{drgrpo}. Early successes are most prominent in mathematical reasoning and code generation, where answers, executions, or tests provide reliable reward signals~\cite{rl4code_SCALEBOX,rlef}. More recent work has extended this paradigm to broader verifiable settings, including formal reasoning~\cite{grpo}, retrieval augmented reasoning~\cite{searchr1}, and agentic tasks with environment feedback~\cite{webagentr1}. 
Complementing reward based optimization, on-policy distillation provides token level teacher feedback on student generated responses~\cite{sopd,think_opd}. \methodshort{} studies this setting with a teacher that shares the student's parameters but observes additional video frames. It uses dense sparse likelihood differences on the same responses to determine selective distillation weights while retaining sparse frame rollouts.

\subsection{Reinforcement Learning for Video Language Reasoning}
Video language models introduce an additional temporal dimension to multimodal reasoning, requiring policy optimization to account for temporally structured and often sparsely localized visual evidence~\cite{tempr1}. Recent video and multimodal RL methods have therefore moved beyond sequence level outcome rewards toward temporal aware objectives, structured policy optimization, and finer grained supervision~\cite{R1VL,R1_SHAREVL,Avatar}. Video-R1 adapts rule-based RL to video reasoning and introduces Temporal GRPO (T-GRPO), which encourages sensitivity to frame order. Video-KTR further performs modality aware policy shaping through key token attribution, while PRPO~\cite{PRPO} study finer grained visual evidence for policy optimization.
Other approaches explore complementary forms of dense supervision, like VISD~\cite{VISD} introduces structured privileged feedback for token-level video reasoning.

Despite these advances, fixed or uniformly sampled frame budgets remain a common design choice in video reasoning pipelines~\cite{efficientvideo}. The auxiliary training signal is therefore typically derived from the sampled visual context itself, rather than from how the policy changes when additional frames become available.
\methodshort{} is complementary to this line of work: it retains the GRPO style video-RL objective, but treats a denser view as additional training evidence and transfers its effect to the sparse view policy through a same response fidelity signal. 

%% file: 3method.tex
\section{Method}
\label{sec:method}

\begin{figure}[t]
    \centering
    \includegraphics[width=0.98\textwidth]{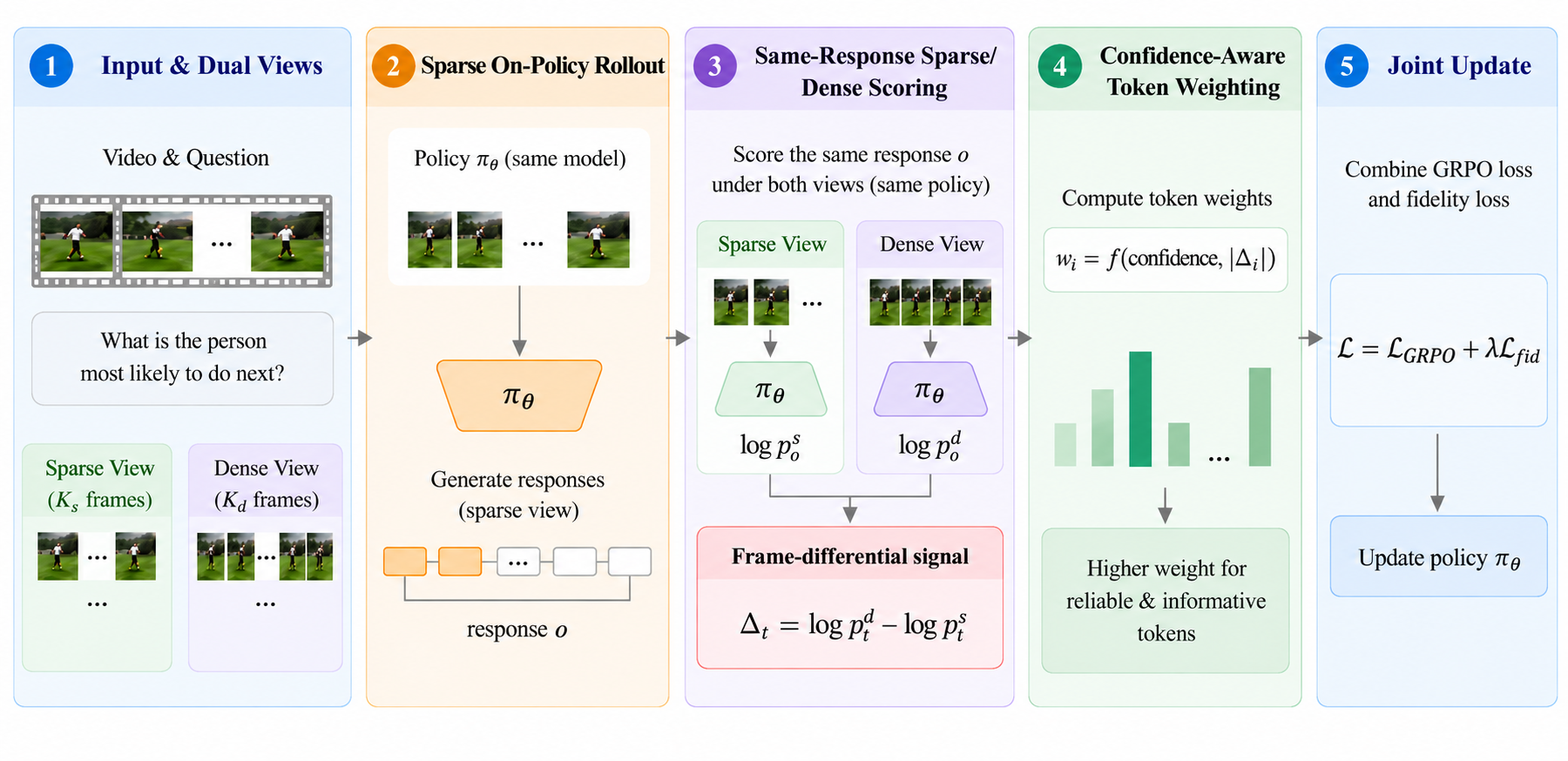}

    \caption{Overview of \methodshort{} for frame differential evidence transfer.
}
    \label{fig:method_overview}
\end{figure}

We propose \textbf{Frame-Differential On-Policy Self-Distillation (\methodshort{})}, which leverages dense frame views to provide richer visual evidence. It transfers this evidence to a policy trained with sparse frame rollouts. As illustrated in Figure~\ref{fig:method_overview},
the policy first generates responses from a sparse view. The same model then
scores these fixed responses under a dense view without gradient updates.
The change in token likelihood between the two views identifies where
additional frames affect the response, We call this difference the Fidelity Advantage (FA). We further use dense view confidence
to control how strongly each token is distilled. \methodshort{} therefore retains
the original GRPO update while adding a token-weighted fidelity objective.

\subsection{GRPO and Reward Design}
\label{sec:grpo_reward}

\paragraph{GRPO objective.}
Given a question $x$, a sparse video view $V_s$, and $G$ responses
$\{y^{(g)}\}_{g=1}^{G}$ sampled from the rollout policy
$\pi_{\theta_{\mathrm{old}}}$, we optimize the GRPO objective~\citep{grpo}:
\begin{equation}
\begin{aligned}
    \mathcal{J}_{\mathrm{GRPO}}(\theta)
    =
    \mathbb{E}
    \left[
    \frac{1}{G}
    \sum_{g=1}^{G}
    \frac{1}{T_g}
    \sum_{i=1}^{T_g}
    \min\left(
        r_{g,i}(\theta)\widehat{A}^{(g)},
        \operatorname{clip}\left(
            r_{g,i}(\theta),
            1-\epsilon_{\mathrm{low}},
            1+\epsilon_{\mathrm{high}}
        \right)\widehat{A}^{(g)}
    \right)
    \right],
    \label{eq:grpo_objective}
\end{aligned}
\end{equation}
where the standard KL regularization with respect to the reference policy is omitted from the displayed objective for clarity. Here, $T_g$ is the response length and
\begin{equation}
    r_{g,i}(\theta)
    =
    \frac{
        \pi_{\theta}
        \left(y_i^{(g)}\mid x,V_s,y_{<i}^{(g)}\right)
    }{
        \pi_{\theta_{\mathrm{old}}}
        \left(y_i^{(g)}\mid x,V_s,y_{<i}^{(g)}\right)
    }
    \label{eq:policy_ratio}
\end{equation}
is the token-level likelihood ratio. For the $G$ responses associated with the same prompt, the scalar group-relative advantage is:
\begin{equation}
    \widehat{A}^{(g)}
    =
    \frac{
        R^{(g)}-\mu_R
    }{
        \sigma_R+\epsilon_A
    },
    \qquad
    \mu_R=\frac{1}{G}\sum_{h=1}^{G}R^{(h)},
    \label{eq:group_advantage}
\end{equation}
where $\sigma_R$ is the standard deviation of the group rewards. The same advantage is assigned to all valid tokens in response $y^{(g)}$.

\paragraph{Reward design.}
The rollout reward combines an answer score with a length reward:
\begin{equation}
    R(y,y^\star)
    =
    R_{\mathrm{acc}}(y,y^\star)
    +
    R_{\mathrm{think}}(y),
    \label{eq:reward}
\end{equation}
where $R_{\mathrm{acc}}\in[0,1]$ is computed by the verifier associated with the question type. Let $L(y)$ denote the length of the valid reasoning span. We define:
\begin{equation}
    R_{\mathrm{think}}(y)
    =
    \begin{cases}
        -0.5, & L(y)<50,\\
        \phantom{-}0.3, & 50\leq L(y)\leq512,\\
        -0.1, & L(y)>512.
    \end{cases}
    \label{eq:thinking_reward}
\end{equation}
A response without a valid reasoning span is assigned to the first case. This term discourages missing or collapsed reasoning as well as excessive verbosity.

\subsection{Frame Differential Self-Teacher}
\label{sec:frame_differential}

For each video $V$, we construct a sparse view $V_s=S(V;m)$ and a dense view $V_d=D(V;M)$, where $m \subset M$. All responses are generated only from the sparse view:
\begin{equation}
    y^{(g)}
    \sim
    \pi_{\theta_{\mathrm{old}}}
    \left(\cdot\mid x,V_s\right).
    \label{eq:sparse_rollout}
\end{equation}
For a sampled response $y=(y_1,\ldots,y_T)$, we record its sparse view token log-probability
\begin{equation}
    \ell_i^s
    =
    \log\pi_{\theta_{\mathrm{old}}}
    \left(y_i\mid x,V_s,y_{<i}\right).
    \label{eq:sparse_logprob}
\end{equation}
We then reuse the same model as a dense self-teacher. Keeping the response fixed, the teacher computes:
\begin{equation}
    \ell_i^d
    =
    \operatorname{sg}\!\left[
        \log\pi_{\theta_{\mathrm{old}}}
        \left(y_i\mid x,V_d,y_{<i}\right)
    \right],
    \label{eq:dense_logprob}
\end{equation}
where $\operatorname{sg}(\cdot)$ denotes stop-gradient. This is a teacher forced scoring pass: the dense self-teacher evaluates the student response but does not generate a second response. The raw frame differential advantage score (raw FA) is the change in sampled token log-probability between the two views:
\begin{equation}
    \Delta_i
    =
    \ell_i^d-\ell_i^s.
    \label{eq:raw_fa}
\end{equation}
A dense view may shift the likelihood of the entire response. To focus on
which positions are affected more strongly than the response wide shift, we
center $\Delta_i$ over valid response tokens:
\begin{equation}
    \widehat{\Delta}_i
    =
    \Delta_i
    -
    \frac{1}{N}
    \sum_{j=1}^{T}m_j\Delta_j,
    \qquad
    N=\sum_{j=1}^{T}m_j,
    \label{eq:centered_fa}
\end{equation}
where $m_i\in\{0,1\}$ is the response mask. Thus,
$\widehat{\Delta}_i>0$ means that the dense view likelihood gain at
position $i$ is above the response level average, while
$\widehat{\Delta}_i<0$ means that it is below the average. The centered
score is used for token weighting; the distillation target remains the
original dense view log-probability $\ell_i^d$.

\subsection{Confidence Aware Selective Distillation}
\label{sec:selective_distillation}

More frames do not guarantee a more reliable prediction: dense views may
also contain redundant or distracting evidence. We therefore use the
confidence of the dense self-teacher to control conservative correction.
Let $\mathcal{K}_i$ contain the indices of the teacher's top-$K_c$ logits
$z_{ij}^d$. We first renormalize their probabilities:
\begin{equation}
    \widetilde{q}_{ij}
    =
    \frac{\exp(z_{ij}^d)}
    {\sum_{k\in\mathcal{K}_i}\exp(z_{ik}^d)},
    \qquad
    j\in\mathcal{K}_i,
    \label{eq:topk_distribution}
\end{equation}
and define confidence using normalized entropy:
\begin{equation}
    c_i
    =
    1-
    \frac{
        -\sum_{j\in\mathcal{K}_i}
        \widetilde{q}_{ij}\log\widetilde{q}_{ij}
    }{
        \log K_c
    }.
    \label{eq:teacher_confidence}
\end{equation}
This score lies in $[0,1]$; a larger value indicates a sharper top-$K_c$ teacher distribution. Suppressing the rollout index for clarity, we assign each response token the fidelity weight:
\begin{equation}
    w_i
    =
    m_i\,\mathbf{1}[R>0]\cdot
    \begin{cases}
        \alpha_{+},
            & \widehat{\Delta}_i>0,\\
        \alpha_{-},
            & \widehat{\Delta}_i<0
              \ \text{and}\ c_i>\tau,\\
        0,
            & \text{otherwise}.
    \end{cases}
    \label{eq:fidelity_weight}
\end{equation}
Tokens with positive relative frame differential scores receive the supportive weight $\alpha_{+}$. For negative scores, the dense prediction is used only when the teacher is confident, and its contribution is reduced by $\alpha_{-}<\alpha_{+}$. This asymmetric design gives full weight to positions with relatively high dense view gains, while applying negative relative changes only conservatively. The reward gate removes non-positive reward trajectories from self-distillation. The centered FA determines token weights, while the current teacher student discrepancy $(d_i)$ determines the local direction of the fidelity update.

\subsection{Joint Training Objective}
\label{sec:joint_objective}

During the policy update, let:
\begin{equation}
    \ell_i^\theta
    =
    \log\pi_\theta
    \left(y_i\mid x,V_s,y_{<i}\right),
    \qquad
    d_i=\ell_i^d-\ell_i^\theta.
    \label{eq:current_student_logprob}
\end{equation}
We use $\phi(d_i)$ to measure the token level discrepancy between the dense self-teacher and the sparse policy.
\begin{equation}
    \phi(d_i)
    =
    \exp(d_i)-d_i-1.
    \label{eq:low_variance_kl}
\end{equation}
For the active token set
$\mathcal{A}=\{i\mid m_i=1,\ w_i>0\}$, the fidelity objective is
\begin{equation}
    \mathcal{L}_{\mathrm{fid}}
    =
    \frac{1}{|\mathcal{A}|}
    \sum_{i\in\mathcal{A}}
    w_i\,\phi(d_i),
    \label{eq:fidelity_loss}
\end{equation}
and is set to zero when $\mathcal{A}$ is empty. 
The final actor loss is:
\begin{equation}
    \mathcal{L}
    =
    \mathcal{L}_{\mathrm{GRPO}}
    +
    \lambda\mathcal{L}_{\mathrm{fid}},
    \label{eq:total_loss}
\end{equation}
where $\mathcal{L}_{\mathrm{GRPO}}$ denotes the standard GRPO loss.
The GRPO term learns from sequence level rewards, while the fidelity term transfers dense view evidence to selected positions in the sparse policy.
\methodshort{} adds one no-gradient dense view scoring pass during training, but does not require dense autoregressive rollouts or a separate teacher model. The dense self-teacher is removed after training, leaving the inference procedure unchanged.

%% file: 4experiment.tex
\section{Experiments}
\subsection{Experimental Setup}

\paragraph{Setup.} We train Qwen2.5-VL-7B-Instruct and Qwen3-VL-4B-Instruct on data derived from the Video-R1-260K\cite{videor1} release. For each example, we sample eight responses and compute the empirical accuracy $a=k/8$. We discard examples with $a\geq0.8$ or $a\leq0.2$, retaining the intermediate difficulty examples and approximately 10K video training instances in total. For training, the sparse student rollout and dense teacher scoring use frame budgets of 8 and 32, respectively. Unless otherwise specified, we use a global batch size of 16, eight responses per prompt, a maximum response length of 2,048 tokens, temperature 1.0, and learning rate $10^{-6}$. We use maximum pixel budgets of $128\times28\times28$ during training and $256\times28\times28$ during evaluation. Further implementation details are provided in the Appendix~\ref{apd:appendix}.

\paragraph{Benchmarks.} We evaluate six complementary video reasoning benchmarks. MVBench\cite{mvbench} measures broad video understanding across actions, interactions, and temporal events, while subtitle free VideoMME \cite{videomme}evaluates general video comprehension under diverse perception and reasoning questions. VideoMMMU\cite{videommmu} emphasizes knowledge intensive multimodal reasoning, and the multiple choice subset of MMVU\cite{mmvu} provides a controlled visual reasoning evaluation. TempCompass\cite{tempcompass} targets temporal relations such as event order and duration, whereas VSI-Bench\cite{vsibench} focuses on spatial relations and scene layout. We report per benchmark scores and the average.

\paragraph{Baselines.} We compare \methodshort{} with two groups of baselines. First, we report published results from representative proprietary and open source Video MLLMs to provide broader context for current video reasoning performance. Second, for controlled comparison, we evaluate both backbones with their corresponding instruction checkpoints, standard GRPO with outcome rewards, Video-R1's T-GRPO, and Video-KTR with key token attribution. The controlled RL baselines use the same filtered training data, rollout count, response length budget, and evaluation protocol whenever direct alignment is possible. All methods are evaluated at 16, 32, and 64 frames.





\subsection{Main Results}

We evaluate  \methodshort{} on six video reasoning benchmarks. Table~\ref{tab:main_results} compares  \methodshort{} with published proprietary and open source Video MLLMs, as well as GRPO, T-GRPO, and Video-KTR trained under our controlled protocol. 

\paragraph{Superior Performance of \methodshort{}.} Under the controlled training protocol, \methodshort{} achieves the best average for both backbones at all three inference frame budgets. On Qwen2.5-VL-7B, \methodshort{} improves over the strongest GRPO-style baseline by 1.0, 0.9, and 1.3 points at 16, 32, and 64 frames, respectively. We observe the same trend on Qwen3-VL-4B, with improvements of 1.1, 0.8, and 1.3 points. These consistent gains show that frame differential supervision is effective across model scales and is not tied to a particular inference budget.

\begin{table*}[!t]
    \centering
    \footnotesize
    \setlength{\tabcolsep}{3.2pt}
    \renewcommand{\arraystretch}{0.9}

    \caption{
        Main results.
         Overall is averaged only when all six benchmark results are available.
    }
    \label{tab:main_results}

    \resizebox{\textwidth}{!}{%
    \begin{tabular}{l c ccc ccc c}
        \toprule

        \multirow{2}{*}{Method}
        & \multirow{2}{*}{\makecell{Frames}}
        & \multicolumn{3}{c}{General Video Understanding}
        & \multicolumn{3}{c}{Fine-Grained Video Reasoning}
        & \multirow{2}{*}{Overall}
        \\

        \cmidrule(lr){3-5}
        \cmidrule(lr){6-8}

        &
        & \makecell{MVBench}
        & \makecell{TempCompass}
        & \makecell{MMVU}
        & \makecell{VideoMMMU}
        & \makecell{VideoMME}
        & \makecell{VSI-Bench}
        &
        \\

        \midrule
        \rowcolor{gray!15}
        \multicolumn{9}{c}{\textbf{Proprietary MLLMs}}
        \\

        \midrule

        \multirow{1}{*}{GPT-4o~\cite{openai2024gpt4o}}
        & -- & 64.6 & 73.8 &  75.4 & 61.2 & 71.9 & 34.0 & 63.5
        \\
        \addlinespace[1pt]


        \multirow{1}{*}{GPT-5}~\cite{openai2025gpt5}
        & -- & 74.1 & 83.3 & 82.6 & 84.6 &  86.7 & 55.0 & 77.7
        \\

        \addlinespace[1pt]
        
        \multirow{1}{*}{Gemini-1.5-Pro}~\cite{gemini15}
        & -- & 60.5 &  67.1 &  71.2 & 53.4 &   75.0 & 45.4 & 62.1
        \\

        \addlinespace[1pt]
        \multirow{1}{*}{Gemini-2.5-Pro}~\cite{gemini25pro}
        & -- & 70.6 &   84.3 &   78.4 &  83.6 &    84.3 & 53.5 & 75.8
        \\

        \midrule

        \rowcolor{gray!15}
        \multicolumn{9}{c}{\textbf{Open Source MLLMs}}
        \\
        \midrule

        \multirow{1}{*}{VILA-1.5-8B}~\cite{vila}
        & 64 & -- & 58.8 &   49.2 &   33.8 &  58.2 &  28.9 & --
        \\
        \addlinespace[1pt]
        \multirow{1}{*}{LLaVA-OV-7B}~\cite{llavaov}
        & 64 &  56.7 &  64.2 &   49.2 &   33.8 &  58.2 &   32.4 & 49.1
        \\
        \addlinespace[1pt]
        \multirow{1}{*}{TW-GRPO}~\cite{twgrpo}
        & 16 &  63.3 &   73.3 &    65.8 &    51.3 &  55.1 &  -- & --
        \\
        \multirow{1}{*}{LongVA-7B}~\cite{longva}
        & -- &  -- &  56.9 &   -- &   23.9 &  52.6 &   29.2 & --
        \\
        \addlinespace[1pt]
        \multirow{1}{*}{Video-RTS}~\cite{videorts}
        &  51.2 &  -- &   -- &    66.4 &    52.7 &  63.0 &  -- & --
        \\
        \addlinespace[1pt]
        \multirow{1}{*}{VideoLLaMA2-7B}~\cite{videollama}
        &  16 &  54.6 &   -- &    44.8 &    -- &  47.9 &  -- & --
        \\
        \addlinespace[1pt]
        \multirow{1}{*}{Kangaroo-8B}~\cite{kangaroo}
        &  -- &  61.1 &   62.5 &    -- &    -- &   56.0 &  -- & --
        \\
        \addlinespace[1pt]
        \multirow{1}{*}{Video-R1-7B}
        &  32 &  65.5 &   73.3 &    64.1 &    50.6 &  59.9 &  31.1 & 57.4
        \\
        \addlinespace[1pt]
        \multirow{1}{*}{Qwen2.5-VL-7B}
        &  32 & 60.9 & 72.6 & 62.1 & 49.3 & 56.9 & 34.6 & 56.1
        \\
        \addlinespace[1pt]
        \multirow{1}{*}{Qwen2.5-VL-7B-165K-SFT}
        &  32 &  61.6 &  69.7 &   62.2 &    51.3 &  55.4 &  32.8 & 55.5
        \\
        \addlinespace[1pt]
        \multirow{1}{*}{Qwen3-VL-4B}
        &  32 & 54.4 & 70.4 & 59.6 & 49.4 & 53.6 & 43.6 & 55.2
        \\
        \addlinespace[1pt]
        \midrule
        

        \rowcolor{gray!15}
        \multicolumn{9}{c}{\textbf{Qwen2.5-VL-7B}}
        \\

        \midrule



        \multirow{3}{*}{GRPO}
        & 16 & 64.6 & 72.5 & 64.5 & 49.4 & 56.0 & 35.9 & 57.2
        \\
        & 32 & 65.0 & 73.3 & 65.1 & 49.6 & 58.4 & 36.7 & 58.0
        \\
        & 64 & 66.2 & \textbf{\underline{73.4}} & \underline{66.1} & 50.4 & 60.6 & 38.6 & 59.2
        \\

        \addlinespace[1pt]

        \multirow{3}{*}{T-GRPO}
        & 16 & 64.2 & 72.6 & 62.4 & 50.0 & 57.1 & 36.1 & 57.1
        \\
        & 32 & 65.5 & 72.9 & 64.1 & 50.5 & 60.5 & 36.0 & 58.3
        \\
        & 64 & 65.9 & 73.1 & 65.1 & 50.7 & \textbf{\underline{62.9}} & 38.7 & \underline{59.4}
        \\

        \addlinespace[1pt]

        \multirow{3}{*}{Video-KTR}
        & 16 & 61.2 & 72.5 & 61.2 & 47.4 & 52.9 & 32.9 & 54.7
        \\
        & 32 & 61.2 & 72.6 & 61.2 & 50.2 & 57.2 & 33.4 & 56.0
        \\
        & 64 & 61.5 & 73.1 & 61.9 & 50.1 & 59.1 & 37.3 & 57.2
        \\

        \addlinespace[1pt]

        \multirow{3}{*}{\textbf{\methodshort{}}}
        & 16 & 64.9 & 72.6 & 65.6 & 49.9 & 56.7 & \underline{39.4} & 58.2
        \\
        & 32 & \underline{66.3} & 72.8 & 65.6 & \underline{51.7} & 59.6 & 38.9 & 59.2
        \\
        & 64 & \textbf{\underline{66.5}} & \underline{73.3} & \textbf{\underline{66.2}} & \textbf{\underline{55.0}} & \underline{62.6} & \textbf{\underline{40.8}} & \textbf{\underline{60.7}}
        \\

        \midrule


        \rowcolor{gray!15}
        \multicolumn{9}{c}{\textbf{Qwen3-VL-4B}}
        \\

        \midrule



        \multirow{3}{*}{GRPO}
        & 16 & 61.6 & 72.4 & 63.8 & 48.3 & 55.5 & 46.4 & 58.0
        \\
        & 32 & 62.6 & 72.2 & 64.4 & 50.0 & 59.2 & 48.8 & 59.5
        \\
        & 64 & 62.7 & 72.4 & 65.1 & 52.7 & 60.5 & 51.0 & 60.7
        \\

        \addlinespace[1pt]

        \multirow{3}{*}{T-GRPO}
        & 16 & 61.4 & 72.9 & 62.7 & 49.7 & 56.3 & 48.4 & 58.6
        \\
        & 32 & 62.3 & 73.0 & 63.8 & 50.8 & 57.9 & 51.1 & 59.8
        \\
        & 64 & 62.6 & 73.0 & 64.6 & 51.8 & 60.6 & \underline{51.6} & \underline{60.7}
        \\

        \addlinespace[1pt]

        \multirow{3}{*}{Video-KTR}
        & 16 & 60.8 & 71.2 & 64.6 & 48.0 & 54.0 & 45.7 & 57.4
        \\
        & 32 & 61.7 & 71.4 & 65.6 & 51.5 & 56.8 & 47.1 & 59.0
        \\
        & 64 & 61.4 & 71.4 & 64.0 & 51.5 & \textbf{\underline{61.4}} & 51.0 & 60.1
        \\

        \addlinespace[1pt]

        \multirow{3}{*}{\textbf{\methodshort{}}}
        & 16 & 61.8 & \underline{74.6} & 65.3 & 49.7 & 57.5 & 49.2 & 59.7
        \\
        & 32 & \underline{63.8} & 74.5 & \underline{65.8} & \underline{52.7} & 57.6 & 49.3 & 60.6
        \\
        & 64 & \textbf{\underline{64.0}} & \textbf{\underline{74.7}} & \textbf{\underline{65.9}} & \textbf{\underline{54.6}} & \underline{61.2} & \textbf{\underline{51.7}} & \textbf{\underline{62.0}}
        \\

        \bottomrule
    \end{tabular}%
    }
\end{table*}


\FloatBarrier
\paragraph{Comparison with Published Video Reasoning Models.} At the matched 32-frame setting, Qwen2.5-VL-7B trained with \methodshort{} outperforms Video-R1 on four of the six benchmarks. Specifically, \methodshort{} improves MVBench from 65.5 to 66.3, MMVU from 64.1 to 65.6, VideoMMMU from 50.6 to 51.7, and VSI-Bench from 31.1 to 38.9. The 7.8 point gain on VSI-Bench is particularly notable, indicating a substantial improvement in spatial video reasoning. Compared with TW-GRPO at 16 frames, \methodshort{} improves MVBench and VideoMME by 1.6 points each, while remaining close on TempCompass and MMVU. 

\paragraph{Robustness across Inference Frame Budgets.}
Increasing the inference frame budget improves all compared methods, as more frames provide additional visual information. We therefore focus not on the absolute improvement from 16 to 64 frames, but on whether the advantage of \methodshort{} persists across different inference budgets. \methodshort{} consistently outperforms the strongest controlled baselines at 16, 32, and 64 frames. At 64 frames, it improves Qwen2.5-VL-7B from 59.4 to 60.7 and Qwen3-VL-4B from 60.7 to 62.0, corresponding to gains of 1.3 and 1.3 points, respectively. Notably, these gains remain even though the dense self-teacher observes at most 32 frames during training. This result indicates that the benefit of frame differential supervision is not restricted to the teacher's training time frame budget and remains effective under higher budget inference.

\subsection{Ablation Studies}
We conduct controlled ablations on Qwen2.5-VL-7B and evaluate every variant with 16 frames. We study four aspects of \methodshort{}: the CASD selection rule, the fidelity coefficient, the construction of the dense self-teacher view, and the use of dense frames for scoring versus autoregressive rollout. Results are shown in Table~\ref{tab:ablation}.

\begin{table*}[!t]
    \centering
    \footnotesize
    \setlength{\tabcolsep}{4.0pt}
    \renewcommand{\arraystretch}{0.95}

    \caption{
        Ablation studies of \methodshort{}. The \textbf{dense} view uniformly samples frames from the video, while the \textbf{local-dense} view adds nearby frames around them. 
    }
    \label{tab:ablation}

    \resizebox{\textwidth}{!}{%
    \begin{tabular}{l l ccc ccc c}
        \toprule

        \multirow{2}{*}{Component}
        & \multirow{2}{*}{Variant}
        & \multicolumn{3}{c}{General Video Understanding}
        & \multicolumn{3}{c}{Fine-Grained Video Reasoning}
        & \multirow{2}{*}{Overall}
        \\

        \cmidrule(lr){3-5}
        \cmidrule(lr){6-8}

        &
        & MVBench
        & TempCompass
        & MMVU
        & VideoMMMU
        & VideoMME
        & VSI-Bench
        &
        \\

        \midrule

        \multirow{2}{*}{CASD}
        & FA positive only
        & 64.1 & 71.2 & 64.5 & \underline{50.2} & 57.1 & 34.2 & 56.8
        \\
        & + positive utility gate
        & 64.6 & 71.2 & 64.0 & 49.2 & \textbf{\underline{58.2}} & 35.4 & 57.1
        \\
        & + answer correctness gate
        & 64.7 & 72.5 & 65.6 & 49.7 & 57.1 & 36.4 & \underline{57.7}
        \\
        & All token distillation
        & 64.6 & 72.3 & 64.5 & 49.7& \underline{57.9} & 35.4 & 57.4
        \\

        \addlinespace[2pt]

        \multirow{3}{*}{Fidelity coefficient}
        & $\lambda=0.005$
        & \underline{64.8} & 72.3 & \textbf{\underline{66.4}} & 49.0 & 56.8 & 33.4 & 57.1
        \\
        & $\lambda=0.05$
        & 63.7 & 71.5 & 63.4 & 47.4 & 57.1 & \underline{36.5} & 56.6
        \\
        & $\lambda=0.1$
        & 20.9 & 45.6 & 43.8 & 39.1 & 29.6 & 20.5 & 33.2
        \\

        \addlinespace[2pt]

        \multirow{3}{*}{Dense self-teacher}
        & 16 frames teacher
        & 64.6 & 71.8 & 64.1 & 48.2 & 56.9 & 35.7 & 56.8
        \\
        & 32 frames, dense view
        & 64.2 & 72.4 & 64.0 & 49.7 & 57.5 & 35.9 & 57.3
        \\
        & 32 frames, local dense view
        & 64.1 & 72.5 & 63.5 & 48.1 & 57.6 & 33.5 & 56.5
        \\

        \addlinespace[2pt]

        Dense frame usage
        & 32 frames dense rollout GRPO
        & 64.6 & \textbf{\underline{72.8}} & 65.4 & \textbf{\underline{50.3}} & 56.6 & 35.9 & 57.6
        \\

        \addlinespace[2pt]

        \textbf{Full method}
        & \textbf{\methodshort{}}
        & \textbf{\underline{64.9}}
        & \underline{72.6}
        & \underline{65.6}
        & 49.9
        & 56.7
        & \textbf{\underline{39.4}}
        & \textbf{\underline{58.2}}
        \\

        \bottomrule
    \end{tabular}%
    }
\end{table*}
\textbf{Effect of Confidence Aware Selection.}
We first examine whether frame differential supervision should be applied indiscriminately. Distilling only positive FA tokens achieves an average score of 56.8; adding the positive utility trajectory gate improves it to 57.1. The complete CASD rule further reaches 58.2, a 1.1 point gain over positive only gated distillation, suggesting that confidence filtered negative FA tokens provide complementary supervision. We also compare against all token distillation, which retains the same trajectory gate but assigns unit weight to every valid token. Its average score of 57.4 is 0.8 points below the full method, supporting the contribution of selective token weighting beyond trajectory filtering. Finally, replacing the positive utility gate with an answer correctness gate decreases performance by 0.5 points, suggesting that final answer correctness is an imperfect proxy for the utility of token level frame differential signals.

\paragraph{Effect of the Fidelity Coefficient.}
We next vary the fidelity coefficient $\lambda$ to control the balance between GRPO and dense view distillation. Reducing $\lambda$ from 0.01 to 0.005 lowers the average from 58.2 to 57.1 indicating that the transferred signal is too weak to improve performance consistently across tasks. Increasing $\lambda$ to 0.05 reduces the average to 56.6, while $\lambda=0.1$ causes a severe collapse to 33.2. We therefore use $\lambda=0.01$ in the full method. The non-monotonic trend confirms that frame differential fidelity should act as an auxiliary objective: insufficient weight limits evidence transfer, whereas excessive weight overwhelms reward driven policy optimization.

\paragraph{Effect of Dense View Construction.}
We compare three teacher view constructions with the same 32-frame budget. The uniformly sampled dense view samples frames across the entire video, while the local dense view samples neighboring frames around those selected by the student. \methodshort{} retains the student selected frames and fills the remaining budget with uniformly sampled frames from the video. The uniform and local dense variants achieve average scores of 57.3 and 56.5, respectively, compared with 58.2 for \methodshort{}. These results suggest that combining the student’s existing observations with broader temporal coverage provides more effective supervision.

\paragraph{Dense Scoring versus Dense Rollout.}
Finally, we compare \methodshort{} with GRPO trained using 32-frame autoregressive rollouts. Dense rollout GRPO generates responses from the dense view and reaches 57.6, whereas \methodshort{} keeps generation sparse, uses dense frames only for scoring, and reaches 58.2.
The higher overall average of \methodshort{} supports fixed response dense scoring as a more favorable accuracy cost trade-off than dense autoregressive rollout.

\section{Mechanism Analysis}
We further analyze the mechanism behind \methodshort{} from two perspectives: computational efficiency and the behavior of the frame differential signal. First, we compare the training cost of \methodshort{} with existing video-RL baselines to understand the overhead introduced by dense view scoring. Second, we investigate whether FA captures meaningful changes caused by additional video evidence, and whether these changes are concentrated on video relevant content rather than generic language tokens.

\subsection{Computational Cost}

\begin{figure}[t]
    \centering

    \begin{subfigure}[t]{0.49\textwidth}
        \centering
        \includegraphics[width=\linewidth]{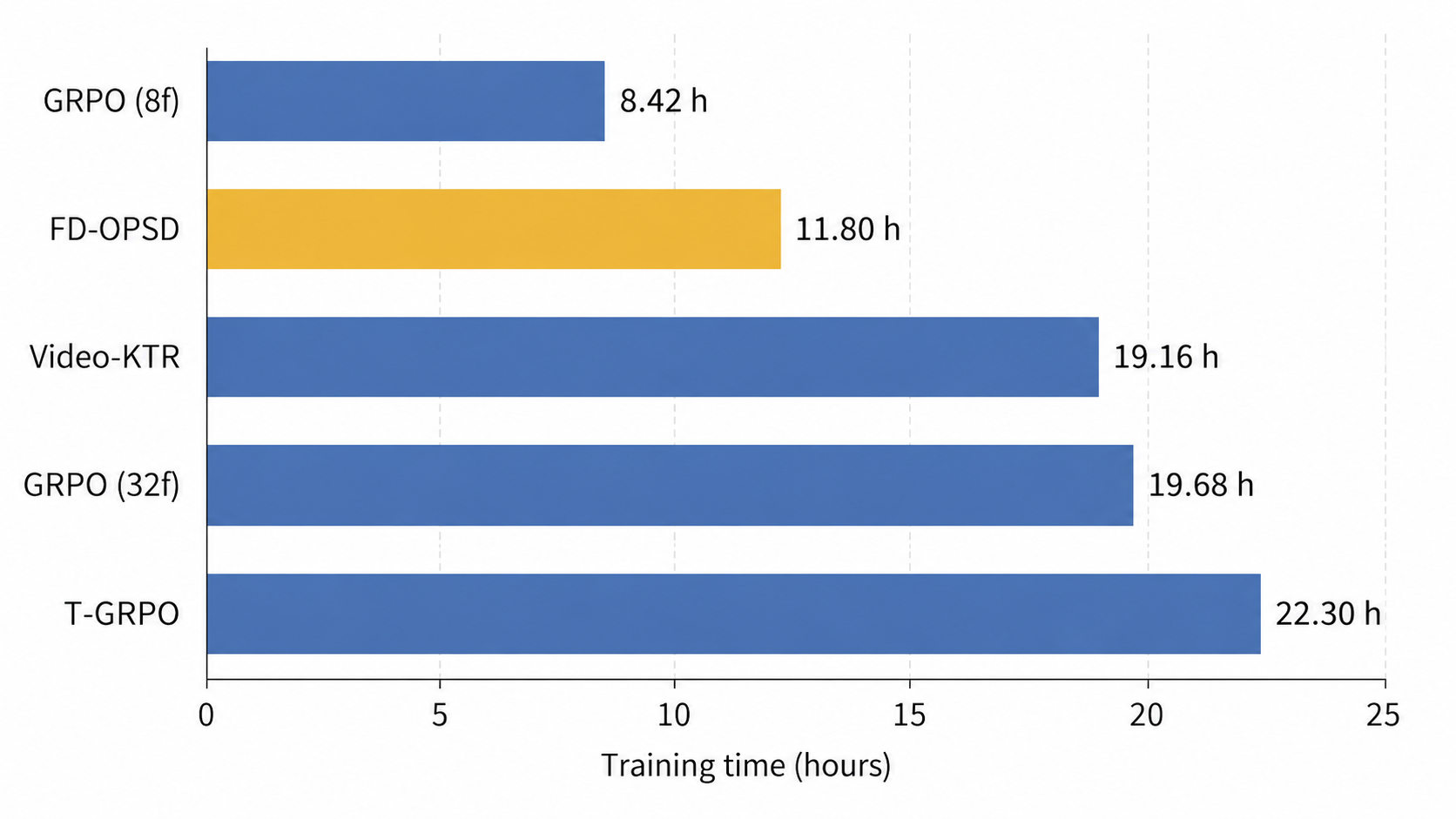}
        \caption{Training time efficiency.}
        \label{fig:cost_time}
    \end{subfigure}
    \hfill
    \begin{subfigure}[t]{0.49\textwidth}
        \centering
        \includegraphics[width=\linewidth]{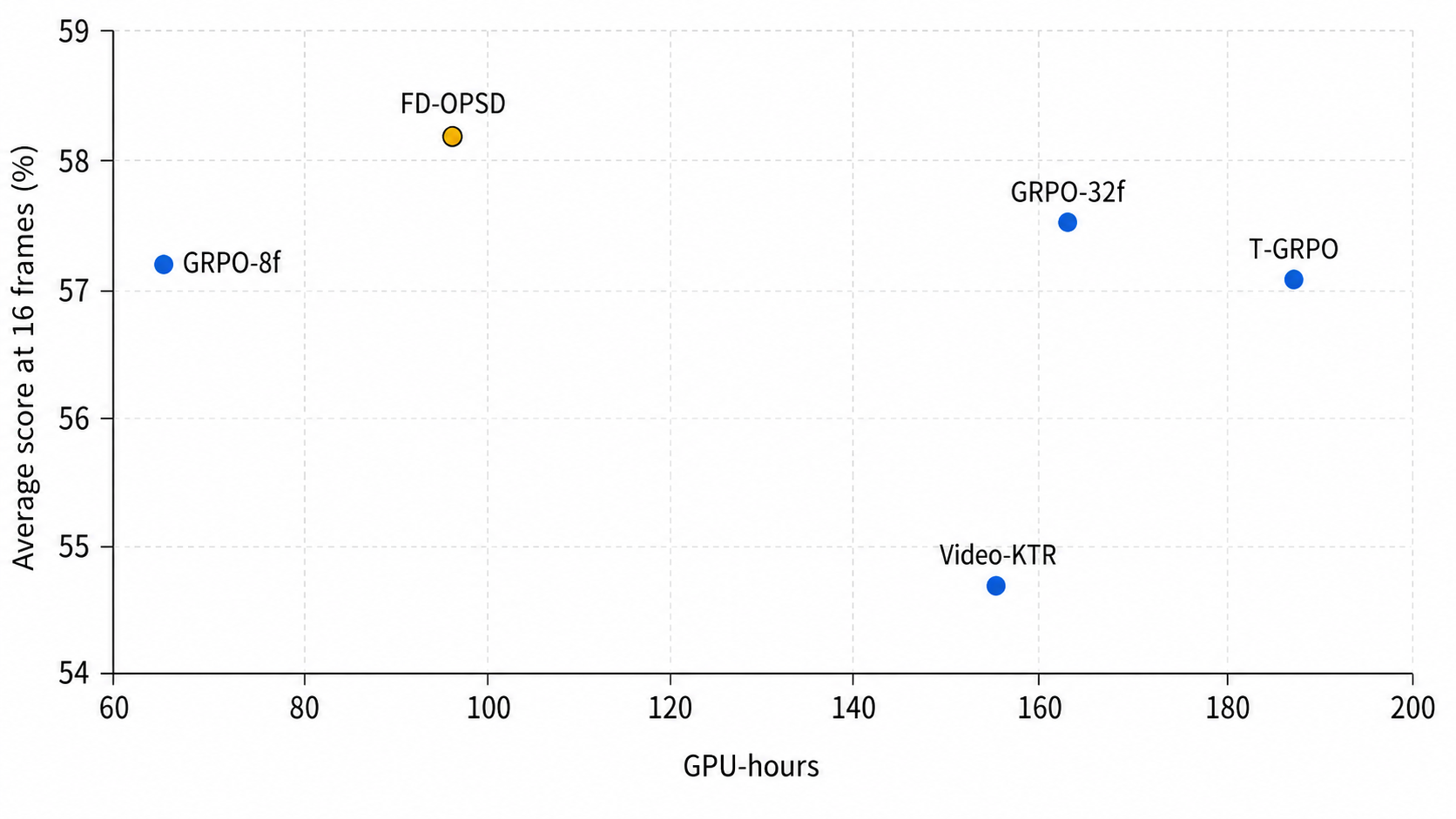}
        \caption{Performance vs. GPU-hours.}
        \label{fig:cost_efficiency}
    \end{subfigure}

    \caption{
        Training cost comparison between \methodshort{} and different video-RL methods. \methodshort{} uses dense frames only for scoring the same on-policy responses, avoiding additional dense autoregressive rollouts.
    }
    \label{fig:computational_cost}
\end{figure}

We compare the end-to-end training cost of \methodshort{} with sparse GRPO, dense rollout GRPO, T-GRPO, and Video-KTR under the same GPUs setup and matched optimization budget, results are shown in Figure~\ref{fig:computational_cost}. Sparse GRPO is the least expensive baseline, requiring 8.42 hours. \methodshort{} completes training in 11.80 hours, a 1.40$\times$ runtime relative to sparse GRPO. 
Despite this overhead, \methodshort{} is substantially more efficient than methods that introduce dense autoregressive rollouts or multiple attribution passes. Compared with 32-frame dense rollout GRPO, \methodshort{} reduces training time by 40.0\% while improving the 16-frame average from 57.6 to 58.2. It also reduces training time by 38.4\% relative to Video-KTR and by 47.1\% relative to T-GRPO, while achieving higher average accuracy. These results show that fixed response dense scoring provides a favorable accuracy cost trade-off: \methodshort{} is more expensive than sparse GRPO, but considerably cheaper than dense rollout and multi-pass video-RL alternatives.

\subsection{CASD Analysis}
We compute category level raw FA at several logged training steps, using 16 prompts at each step, to examine whether the frame differential signal consistently focuses on video relevant content throughout training.
\begin{figure}[t]
    \centering

    \begin{subfigure}[t]{0.49\textwidth}
        \centering
        \includegraphics[width=\linewidth]{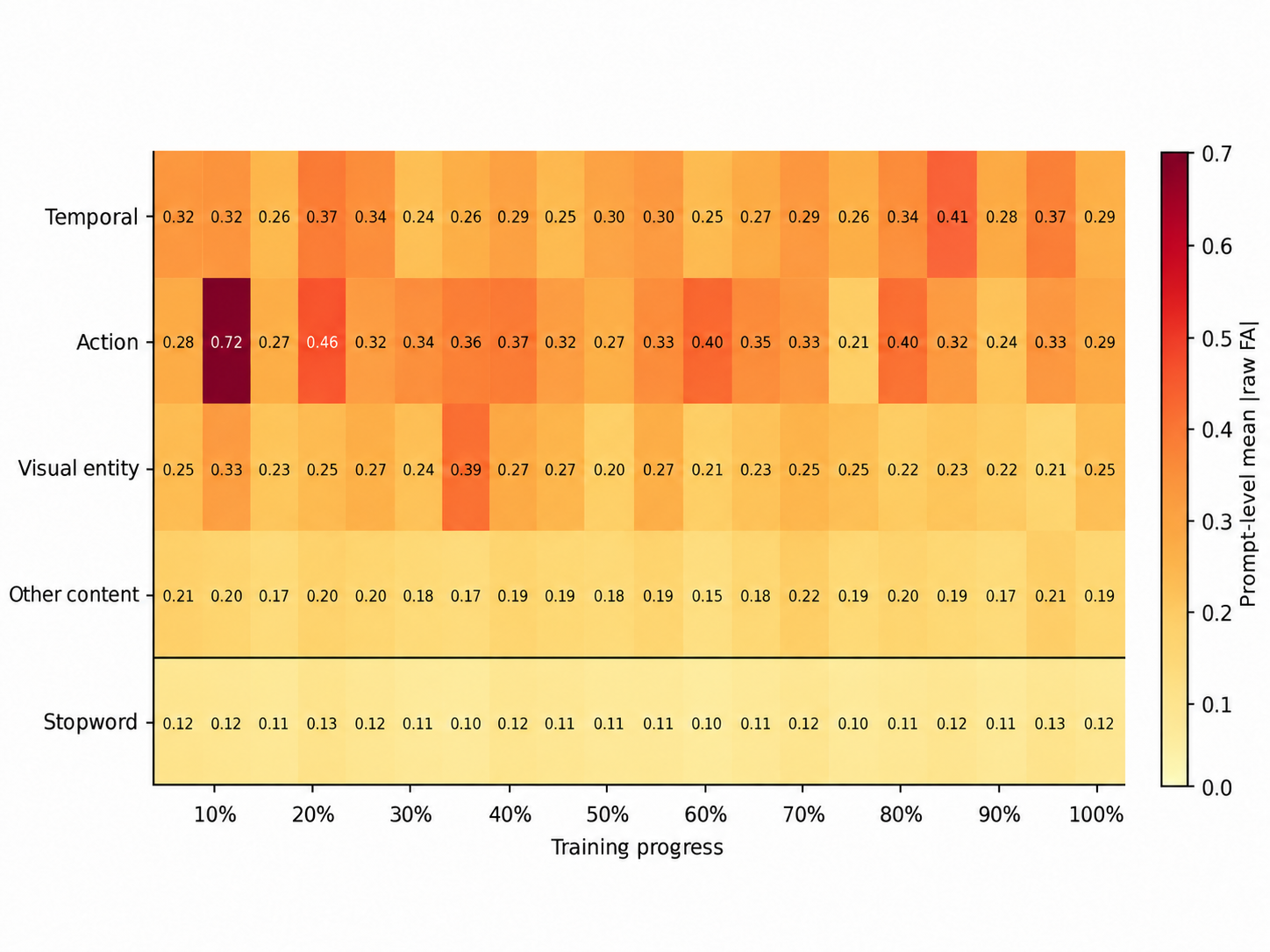}
        \caption{ Category wise prompt level mean absolute raw FA.}
        \label{fig:casd_analysis-a}
    \end{subfigure}
    \hfill
    \begin{subfigure}[t]{0.49\textwidth}
        \centering
        \includegraphics[width=\linewidth]{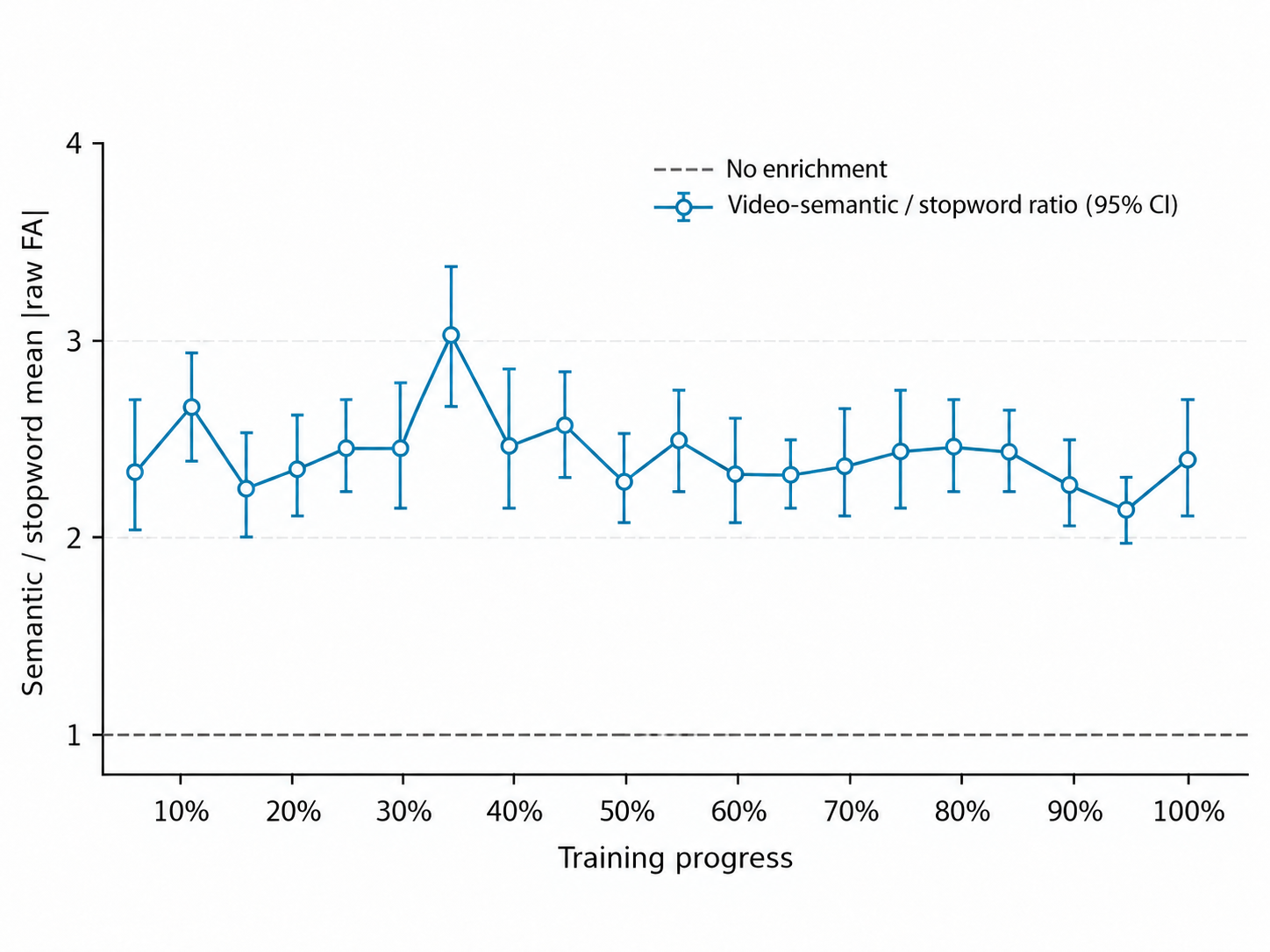}
        \caption{Video semantic enrichment across training.}
        \label{fig:casd_analysis-b}
    \end{subfigure}

    \caption{
Analysis of the frame differential signal throughout training.
(a) Category wise prompt level mean absolute raw FA. 
(b) Video semantic to stopword enrichment across training.    
}
    \label{fig:casd_analysis}
\end{figure}

\paragraph{Frame sensitivity across token categories.}
Figure~\ref{fig:casd_analysis-a} tracks the absolute raw FA of different token categories throughout training. Raw FA measures the change in the log-probability of the sampled token when the same response is scored under the sparse and dense video views. We use its absolute value to measure how strongly each token is affected by the additional frames. Action, temporal, and visual entity tokens have mean absolute raw FA values of 0.35, 0.30, and 0.25, respectively, compared with 0.11 for stopwords. This separation shows that the sparse to dense intervention has a larger effect on video related tokens than on generic function words.

\paragraph{Video-semantic enrichment across training.}
Figure~\ref{fig:casd_analysis-b} further summarizes this separation by comparing the prompt level mean absolute raw FA of video semantic tokens (temporal, action, and visual entity tokens) with that of stopwords. A ratio above one indicates that additional frames affect video semantic content more strongly than generic language tokens. The semantic to stopword ratio averages 2.31, ranges from 1.95 to 3.05, and has a 95\% confidence interval lower bound above one at every logged training step. This persistent separation shows that the frame differential signal remains concentrated on video related content throughout training.


%% file: 5conclusion.tex
\section{Conclusion}
We presented \methodshort{}, which turns the difference between sparse and dense video views into token level supervision for video RL. By scoring the same on policy response under both views, \methodshort{} transfers useful dense evidence without a dense autoregressive rollout or a separate teacher model. Across two Qwen-VL backbones and six benchmarks, the method consistently improves matched GRPO-style baselines while requiring substantially less training time than dense rollout and multi-pass alternatives. The analysis shows that FA is concentrated on video semantic tokens and that CASD filters this signal at both the trajectory and token levels. These results establish frame budget intervention as a practical way to improve sparse rollout video reasoning without changing inference.

%% file: 6limilation.tex
\section{Limitation}
Our evaluation currently covers two Qwen-VL model families and approximately 10K filtered training examples; validation on larger models, higher frame budgets, other architectures, and more diverse video data remains future work. We study three simple dense frame sampling strategies, but do not explore learned or question adaptive frame selection. The self-teacher can only expose knowledge already present in the online policy, and its top-$K$ confidence estimate may be imperfect when the dense view is ambiguous. Future work can scale \methodshort{} across model sizes and frame budgets, learn adaptive dense views, and improve confidence calibration.

\section{AI Use Statement}
Generative AI tools were used for language polishing. The authors take full responsibility for the final content.

%% file: appendix.tex
\section{More Details}
\label{apd:appendix}

\subsection{Implementation Details}
For both models, we selected about 10k problems from the Video-R1-260K datasets. For each problem, we sample 8 times. Both models are trained with 8$\times$80 GPUs. The other hyperparameters used in the training process are presented in the Table~\ref{tab:appendix_implementation}.

\begin{table}[H]
\centering
\caption{Training and evaluation setup for the two model families.}
\label{tab:appendix_implementation}
\resizebox{\linewidth}{!}{
    \begin{tabular}{lcc}
        \toprule
        \textbf{Parameter} 
        & \textbf{Qwen2.5-VL-7B-Instruct} 
        & \textbf{Qwen3-VL-4B-Instruct} \\
        \midrule

        Sparse student frame budget 
        & 8 
        & 8 \\
        
        Dense teacher frame budget 
        & 32 
        & 32 \\
        
        Global batch size 
        & 16 
        & 16 \\
        
        Responses per prompt 
        & 8 
        & 8 \\
        
        Maximum response length 
        & 2,048 
        & 2,048 \\
        
        Temperature 
        & 1.0 
        & 1.0 \\
        Warmup steps 
        & 100 
        & 100 \\
        Epoch 
        & 1.0 
        & 1.0 \\
        
        $\lambda$ Fidelity
        & 0.01 
        & 0.01 \\
        
        Learning rate 
        & 1$\times10^{-6}$ 
        & 1$\times10^{-6}$ \\
        
        Training maximum pixel budget 
        & 128$\times$28$\times$28 
        & 128$\times$28$\times$28 \\
        
        Evaluation maximum pixel budget 
        & 256$\times$28$\times$28
        & 256$\times$28$\times$28 \\
        
        $\alpha_{+}$
        & 1.0
        & 1.0 \\

        $\alpha_{-}$
        & 0.3
        & 0.3 \\
        $K_{c}$
        & 10
        & 10 \\
        $\tau$
        & 0.3
        & 0.3 \\
        
        \bottomrule
        
    \end{tabular}
}
\end{table}

\subsection{Performance Details}
We provide more detailed evaluation results (Table~\ref{tab:aditional_main_results}) beyond the main table, including results for the two Qwen base models and a more comprehensive evaluation of Video-R1-7B.
\begin{table}[H]
    \centering
    \footnotesize
    \setlength{\tabcolsep}{3.2pt}
    \renewcommand{\arraystretch}{0.9}

    \caption{
        Detailed evaluation results beyond the main comparison.
    }
    \label{tab:aditional_main_results}

    \resizebox{\textwidth}{!}{%
    \begin{tabular}{l c ccc ccc c}
        \toprule

        \multirow{2}{*}{Method}
        & \multirow{2}{*}{\makecell{Frames}}
        & \multicolumn{3}{c}{General Video Understanding}
        & \multicolumn{3}{c}{Fine-Grained Video Reasoning}
        & \multirow{2}{*}{Overall}
        \\

        \cmidrule(lr){3-5}
        \cmidrule(lr){6-8}

        &
        & \makecell{MVBench}
        & \makecell{TempCompass}
        & \makecell{MMVU}
        & \makecell{VideoMMMU}
        & \makecell{VideoMME}
        & \makecell{VSI-Bench}
        &
        \\

        \midrule

        \rowcolor{gray!15}
        \multicolumn{9}{c}{\textbf{Open Source MLLMs}}
        \\
        \midrule

        \multirow{3}{*}{Qwen2.5-VL-7B-165K-SFT}
        &  16 &  60.9 &  69.0 &   60.2 &    48.4 &  53.1 &  30.6 & 53.7
        \\
        &  32 &  61.6 &  69.7 &   62.2 &    51.3 &  55.4 &  32.8 & 55.5
        \\
        &  64 &  61.7 &  70.0 &   61.9 &    51.1 &  58.9 &  31.0 & 55.8
        \\
        \addlinespace[1pt]

        \multirow{3}{*}{Video-R1-7B}
        &  16 &  64.2 &   73.2 &    64.6 &    50.2 &  57.8 &  31.0 & 56.8
        \\
        &  32 &  65.5 &   73.3 &    64.1 &    50.6 &  59.9 &  31.1 & 57.4
        \\
        &  64 &  66.1 &   \textbf{\underline{73.4}} &    64.9 &    51.7 &  \underline{61.7} &  32.6 & 58.4
        \\
        \addlinespace[1pt]

        \midrule
        

        \rowcolor{gray!15}
        \multicolumn{9}{c}{\textbf{Qwen2.5-VL-7B}}
        \\

        \midrule

        \multirow{3}{*}{Instruction}
        & 16 & 59.9 & 72.2 & 62.4 & 48.6 & 53.1 & 33.6 & 55.0
        \\
        & 32 & 60.9 & 72.6 & 62.1 & 49.3 & 56.9 & 34.6 & 56.1
        \\
        & 64 & 61.5 & 73.0 & 62.5 & 50.6 & 59.4 & 37.8 & 57.5
        \\

        \addlinespace[1pt]

        \multirow{3}{*}{\textbf{\methodshort{}}}
        & 16 & 64.9 & 72.6 & 65.6 & 49.9 & 56.7 & \underline{39.4} & 58.2
        \\
        & 32 & \underline{66.3} & 72.8 & \underline{65.6} & \underline{51.7} & 59.6 & 38.9 & \underline{59.2}
        \\
        & 64 & \textbf{\underline{66.5}} & \underline{73.3} & \textbf{\underline{66.2}} & \textbf{\underline{55.0}} & \textbf{\underline{62.6}} & \textbf{\underline{40.8}} & \textbf{\underline{60.7}}
        \\

        \midrule


        \rowcolor{gray!15}
        \multicolumn{9}{c}{\textbf{Qwen3-VL-4B}}
        \\

        \midrule

        \multirow{3}{*}{Instruction}
        & 16 & 52.9 & 70.2 & 60.4 & 45.8 & 50.3 & 41.0 & 53.4
        \\
        & 32 & 54.4 & 70.4 & 59.6 & 49.4 & 53.6 & 43.6 & 55.2
        \\
        & 64 & 55.6 & 70.0 & 63.6 & 51.8 & 57.0 & 45.3 & 57.2
        \\

        \addlinespace[1pt]

        \multirow{3}{*}{\textbf{\methodshort{}}}
        & 16 & 61.8 & \underline{74.6} & 65.3 & 49.7 & 57.5 & 49.2 & 59.7
        \\
        & 32 & \underline{63.8} & 74.5 & \underline{65.8} & \underline{52.7} & \underline{57.6} & \underline{49.3} & \underline{60.6}
        \\
        & 64 & \textbf{\underline{64.0}} & \textbf{\underline{74.7}} & \textbf{\underline{65.9}} & \textbf{\underline{54.6}} & \underline{61.2} & \textbf{\underline{51.7}} & \textbf{\underline{62.0}}
        \\

        \bottomrule
    \end{tabular}%
    }
\end{table}
\FloatBarrier

\section{Deeper Analysis} 
\subsection{CASD separates support, correction, and rejection.}
We visualize three representative responses selected with a fixed quantile based rule, covering moderate to strong semantic contrast without manual cherry-picking. In the two positive reward cases, the dense view supports some video related tokens while assigning lower likelihood to others. For example, it favors the action token ``cutting'' but reduces the likelihood of ``container.'' 
These changes illustrate frame dependent token preferences, which CASD selectively distills using response centered FA and teacher confidence.
The third case contains strong visual and temporal FA but has negative total reward. The trajectory gate therefore removes the entire fidelity update. Together, the cases show that FA identifies where additional frames change the prediction, while reward and confidence determine whether and how that change is transferred to the sparse policy.
\begin{figure}[H]
    \centering

    \includegraphics[width=0.98\textwidth]{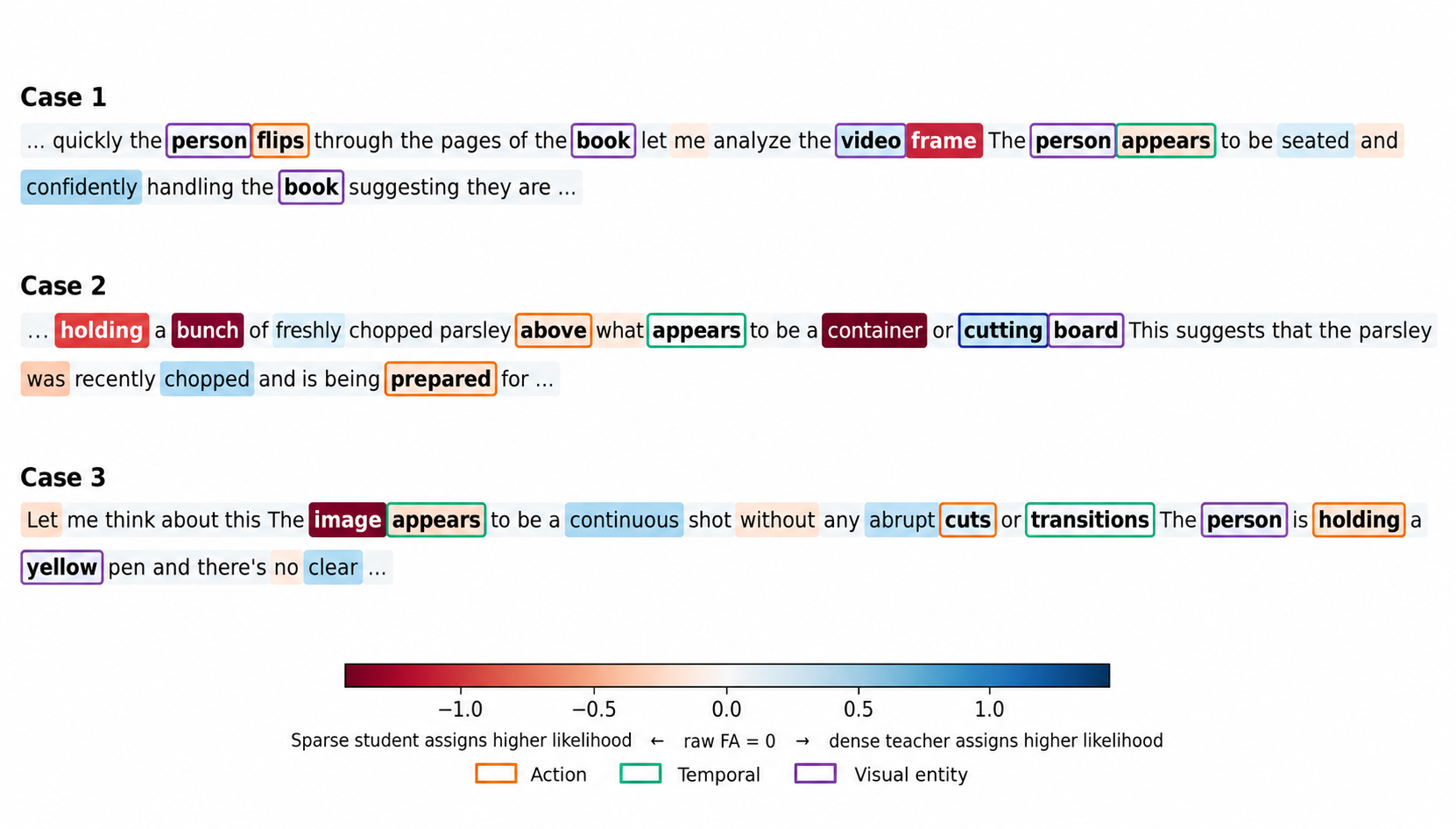}
    \caption{
    Representative examples of CASD support, correction, and rejection under sparse and dense video views.
}
    \label{fig:case-sd}
\end{figure}

\subsection{When Does Additional Frame Evidence Help?}
We first test whether a larger frame budget provides useful evidence. We random smaple 500 questions from test datasets. On the same 500 test questions, \textbf{uniform-4} uses four evenly spaced frames, \textbf{random-4} and \textbf{random-8} sample four or eight frames at random, and \textbf{dense-16} uses up to sixteen frames spread over the video. All variants use the same Qwen2.5-VL-7B checkpoint and results are in Table~\ref{tab:frame_sampling}.

\begin{table}[H]
\centering
\caption{Effect of frame sampling on answer accuracy. 
}
\label{tab:frame_sampling}
\small
\begin{tabular}{lccc}
\toprule
\textbf{Sampling Strategy} 
& \textbf{Accuracy (\%)} 
& \textbf{$\Delta$ vs. Dense-16} \\
\midrule
Uniform-4     & 42.8 & $-5.8$ & \\
Random-4 (1)  & 44.2 & $-4.4$ & \\
Random-4 (2)  & 44.8 & $-3.8$ & \\
Random-8      & 46.2 & $-2.4$ & \\
\textbf{Dense-16} 
              & \textbf{48.6} & -- & \\
\midrule
\multicolumn{4}{l}{
\footnotesize Sparse correct / Dense-16 wrong: about 8.0\% of questions.
} \\
\bottomrule
\end{tabular}
\end{table}

Dense-16 reaches 48.6\% accuracy, compared with 42.8\% for uniform-4, 44.2\% and 44.8\% for two random-4 draws, and 46.2\% for random-8. We also compare the paired predictions on each question by counting how often one sampler succeeds while the other fails. The difference is clearest for dense-16 versus uniform-4.
More frames are helpful on average but are not always better. A sparse sampler succeeds while dense-16 fails on 8.0\% of the questions. This is the setting FD-OPSD is designed for: the dense view provides extra evidence, while the reward and confidence gates prevent the policy from treating every dense-view change as a reliable target.

\paragraph{The benefit is example dependent.}
More frames do not improve every prediction. In 8.0\% of test examples, a sparse strategy answers correctly while dense-16 fails. We therefore use the dense view as a source of conditional supervision rather than treating it as an oracle. This observation motivates the two safeguards in CASD: we prioritize positions with positive centered FA and include those with negative centered FA only when the dense self-teacher is confident.

\subsection{Where Does the Frame Differential Signal Concentrate?}

\begin{figure}[H]
    \centering

    \begin{subfigure}[t]{0.49\textwidth}
        \centering
        \includegraphics[width=\linewidth]{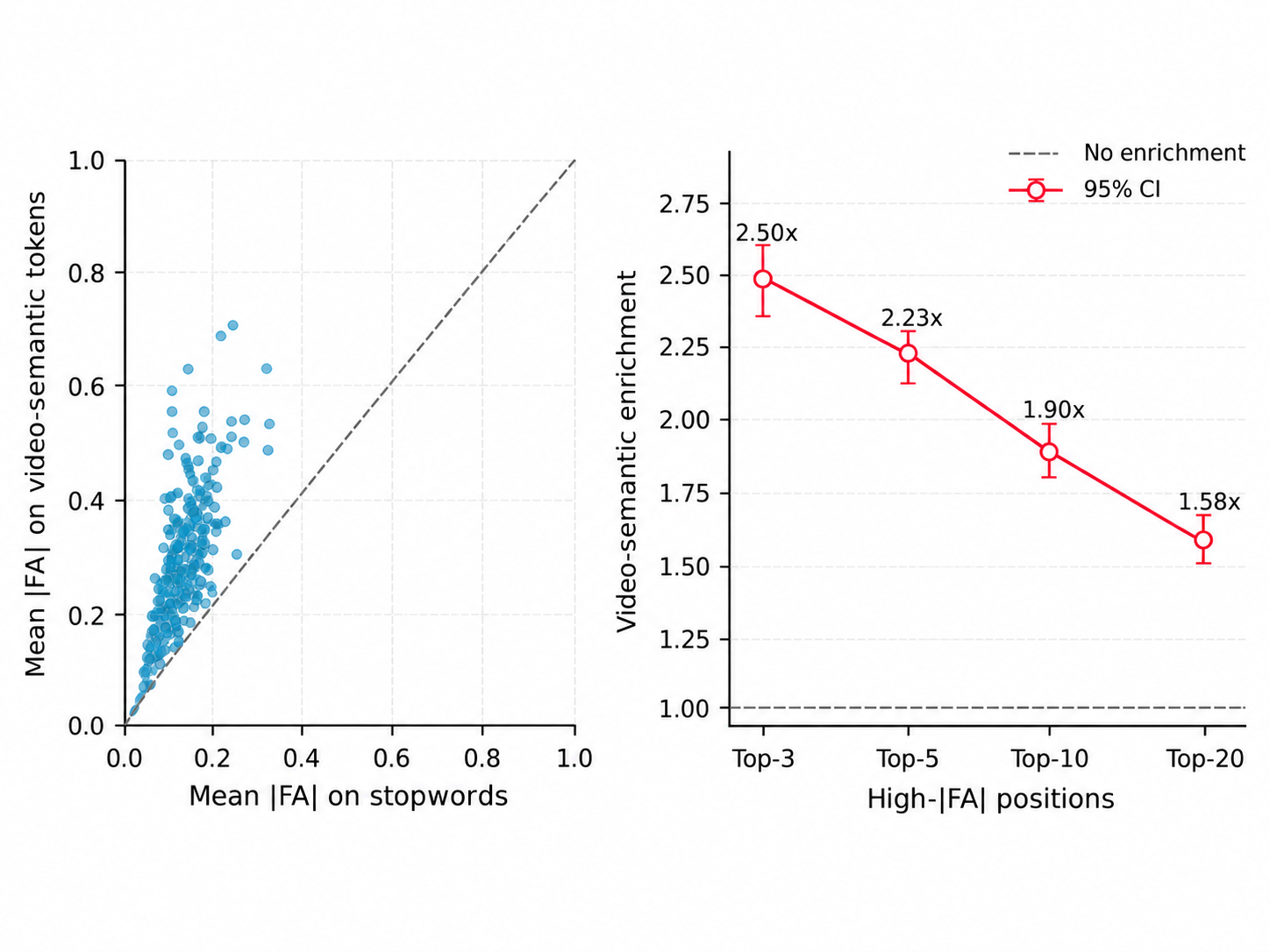}
        \caption{Response centered FA.}
        \label{fig:case}
    \end{subfigure}
    \hfill
    \begin{subfigure}[t]{0.49\textwidth}
        \centering
        \includegraphics[width=\linewidth]{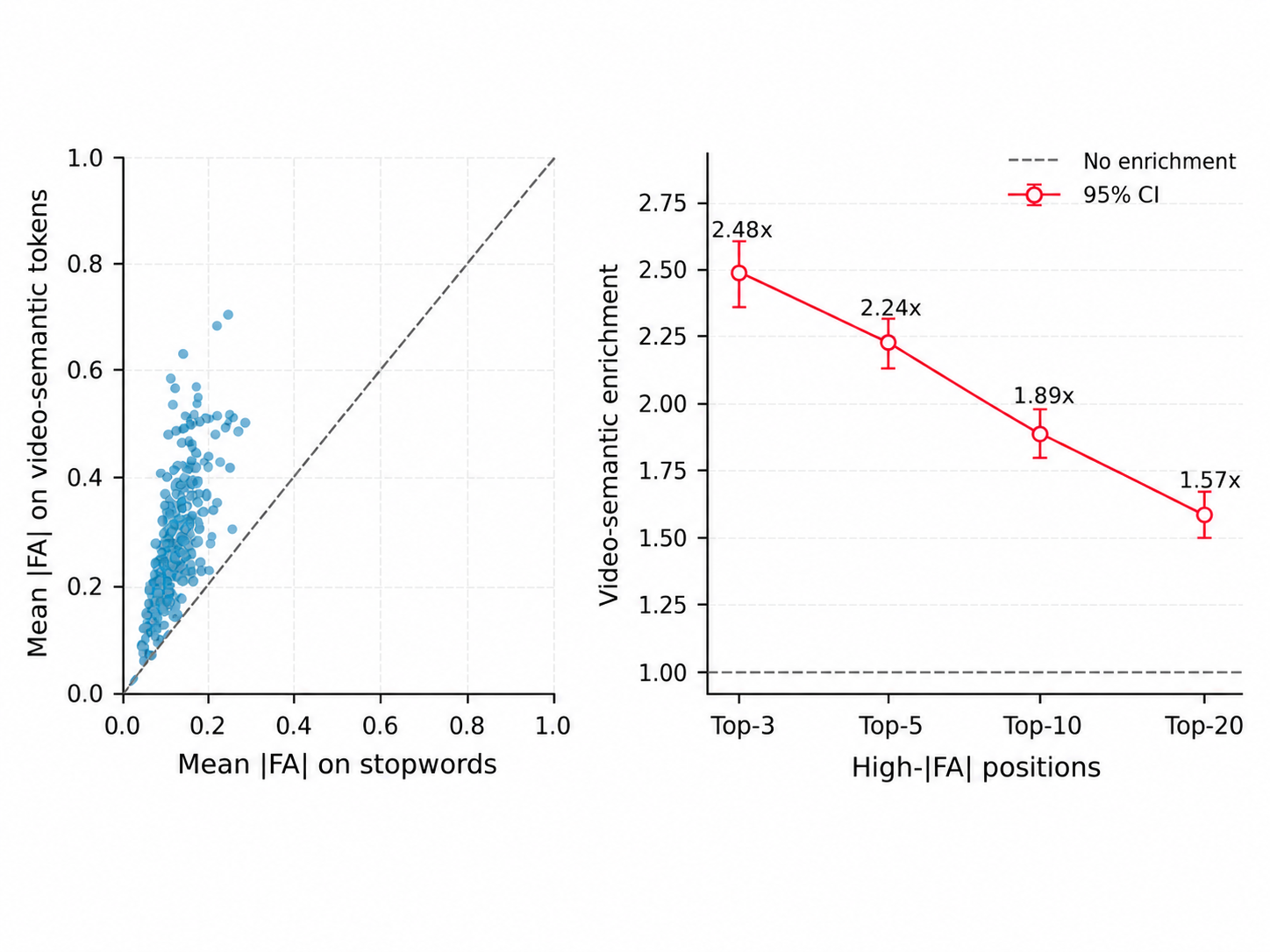}
        \caption{Raw FA.}
        \label{fig:training}
    \end{subfigure}

    \caption{
     Video semantic concentration of the frame differential signal. Each panel compares the prompt level mean absolute FA of video semantic
tokens and stopwords, and reports video semantic enrichment among high FA positions.
}
    \label{fig:fa_analysis}
\end{figure}

\paragraph{Frame differential changes concentrate on video semantic tokens.}
We group temporal, action, and visual entity tokens as video semantic tokens and compare their absolute FA with that of stopwords. As shown in Figure~\ref{fig:case}, video semantic tokens exhibit
larger response centered FA than stopwords, with a  video  semantic to stopword ratio of 2.03$\times$ [1.96, 2.10]. Figure~\ref{fig:training} shows a similar ratio of 2.30$\times$ [2.22, 2.40] for raw FA, indicating that this concentration is already present before response level centering.

\paragraph{High FA positions show stronger video semantic enrichment.}
We further examine the positions with the largest absolute FA. Video semantic enrichment reaches 2.50$\times$ for the top three positions, 2.23$\times$ for the top five, 1.90$\times$ for the top ten, and 1.58$\times$ for the top twenty. The stronger enrichment among high FA positions indicates that the frame differential signal preferentially highlights tokens whose predictions are more sensitive to additional video evidence. This supports the use of FA for selective token level distillation in \methodshort{}.